\pdfoutput=1

\documentclass[11pt]{article}

\usepackage{emnlp2021}

\usepackage{times}
\usepackage{latexsym}
\usepackage{graphicx}
\usepackage{hhline}
\usepackage{chngcntr}
\usepackage{multirow}
\usepackage{comment}
\usepackage{diagbox}
\usepackage{amsmath}
\usepackage{listings}
\DeclareMathOperator*{\argmin}{argmin}
\usepackage[T1]{fontenc}
\usepackage[flushleft]{threeparttable}
\usepackage[utf8]{inputenc}
\usepackage{microtype}

\title{Three-level Hierarchical Transformer Networks for \\ Long-sequence and Multiple Clinical Documents Classification}
\author{Yuqi Si \and Kirk Roberts \\
        School of Biomedical Informatics, \\The University of Texas Health Science Center, Houston, USA \\
        \texttt{\{yuqi.si, kirk.roberts\}@uth.tmc.edu}}

\begin{document}
\maketitle
\begin{abstract}
We present a Hierarchical Transformer Network for modeling long-term dependencies across clinical notes for the purpose of patient-level prediction.
The network is equipped with three levels of Transformer-based encoders to learn progressively from words to sentences, sentences to notes, and finally notes to patients.
The first level from word to sentence directly applies a pre-trained BERT model as a fully trainable component. While the second and third levels both implement a stack of transformer-based encoders, before the final patient representation is fed into a classification layer for clinical predictions.
Compared to conventional BERT models, our model increases the maximum input length from 512 tokens to much longer sequences that are appropriate for modeling large numbers of clinical notes.
We empirically examine different hyper-parameters to identify an optimal trade-off given computational resource limits. Our experiment results on the MIMIC-III dataset for different prediction tasks demonstrate that the proposed Hierarchical Transformer Network outperforms previous state-of-the-art models, including but not limited to \textsc{BigBird}. 

\end{abstract}

\section{Introduction}

Transformers have gained popularity and have achieved superior performance in many natural language processing (NLP) tasks. The scheme of Transformers entirely dispenses with convolution and recurrence, solely relying on multi-headed self-attention mechanisms and position-wise feed forward networks \cite{vaswani2017attention}. Inspired by Transformers, the BERT model \cite{devlin2019bert} and its variants \cite{lan2019albert,liu2019roberta,sanh2019distilbert,joshi2020spanbert,zaheer2020big} have been solidly established as the state-of-the-art methods in numerous NLP studies. BERT-based models impose an input length constraint, which limits their applicability of processing multiple, longitudinal documents.
To handle this challenge, previous efforts have proposed to split long documents (or, by extension, a sequence of documents) into small chunks and then aggregate their respective representations \cite{adhikari_docbert_2019, pappagari_hierarchical_2019}.
However, these approaches do not consider the temporal interrelations between longitudinal sequences of (potentially many) documents, and also disregard the knowledge of hierarchical structure within the document \cite{yang_beyond_2020}.
For humans, it is important to understand hierarchical and longitudinal document structure when reading a series of long documents, such as chapters in a full-length novel, legal documents, and clinical notes in patient trajectories. Similarly, to process longitudinal documents, a model should incorporate this information into its architecture.

Motivated by Hierarchical Attention Networks \cite{yang2016hierarchical}, we propose Hierarchical Transformer Networks to capture the structure inherent in longitudinal sequences of documents.
Our model constructs three levels---from words to sentences, then sentences to documents, and finally documents to the prediction label---leveraging both temporal and structural interrelations.
We utilize a BERT model directly at the word level, experimenting with different sized BERT models to evaluate the relative trade-off between model size and sequence length.
At the sentence and document levels, we employ a Transformer-based encoder architecture first proposed in \citet{vaswani2017attention}. 
Also, we implement a time-aware adaptive segmentation at the document level to capture the real temporal relationship of notes across long time periods, while aggregating notes in short time periods. 

We conduct experiments using clinical notes from MIMIC-III \cite{johnson2016mimic}.
Due to the difficulty of training Transformers successfully \cite{popel_training_2018}, we extensively experiment with numerous hyper-parameter settings to achieve a robust training system.
We also integrate distributed training to resolve memory constraints and to incorporate longer input texts. 
We compare our proposed model with the state-of-the-art models for two clinical outcome predictions: in-hospital mortality and phenotype prediction. 
Our experimental evaluation shows that Hierarchical Transformer Networks consistently outperform other alternatives with an overall improvement of up to 21\% in AUC, 51\% in PRC and 46\% in F1 score. 
Through extensive ablation studies, we show that the components of the Hierarchical Transformer Networks successfully process temporal and hierarchical information of clinical notes and effectively enhance clinical predictions.

\section{Related Work}

\subsection{Hierarchical Deep Learning Architecture}

To handle long documents, previous works have applied hierarchical deep learning models that stack neural networks to draw inference at each level of the hierarchy \cite{zhou2016attention,gao2018hierarchical}.
\citet{yang2016hierarchical} first proposed the hierarchical attention network based on GRUs for document classification.
\citet{kowsari2017hdltex} later applied multiple deep learning architectures, including fully-connected DNN, GRU, LSTM, and CNN into a hierarchical model.
More recent work, HiBERT \cite{zhang_hibert_2019}, presented a hierarchical architecture to pre-train document-level Transformer encoders with unlabeled data for extractive document summarization.
These hierarchical models progressively learn a representation for long-term dependencies, which could in theory enable them to explicitly deal with longitudinal sequences of documents.

\subsection{Transformer Models in Clinical Domain}

With the wide implementation of Transformer-based models in NLP, these have also been adapted to clinical tasks.
One category of such tasks is clinical predictive modeling \cite{SI2021103671}.
In a similar paradigm of sequence modeling with Recurrent Neural Networks, Transformers attempt to model the entire patient trajectory by encoding clinical events at each time stamp\cite{choi2020learning}. 
One of the earliest efforts intended to develop a multi-headed attention-based model for processing multivariate clinical time series data \cite{song2018attend}. Recently, BEHRT \cite{li2020behrt} was built based on BERT for analyzing large-scale, sequential clinical data. 
BEHRT integrated patients’ previous diagnoses and demographic information to predict future diagnoses. 
However, none of these works incorporated text data.
Another notable domain where Transformers continue to push the frontier is clinical NLP. Many studies pre-trained BERT models with biomedical literature \cite{lee2020biobert,beltagy2019scibert} or clinical notes \cite{alsentzer2019publicly,peng2019transfer,si2019enhancing} to develop the domain-specific language model, and these studies showed that such models generally outperform off-the-shelf models in varied clinical NLP tasks.
However, for clinical text classification (e.g., automatic ICD coding, clinical outcome predictions) which generally requires a series of clinical notes as input, BERT does not always perform well, probably due to its restriction on computational resources and its fixed-length restriction \cite{Li_Yu_2020, makarenkov2020lessons, si2020patient}.
In keeping more closely with the spirit of Transformers, our work is also built on top of Transformers with an emphasized focus on effective representation of long document sequences.

\subsection{Clinical Text Classification}
Unstructured notes contain important details about patient status that do not exist in the structured data of Electronic Health Records (EHR). 
Previous studies have developed advanced neural networks to classify clinical notes with word embeddings \cite{pmlr-v85-liu18b}. Despite the success, context-free word embeddings fail to encode the information of a given surrounding context \cite{si2019enhancing}. 
More advanced pre-trained language models show their capability to provide context-sensitive representations for clinical words \cite{feng-etal-2020-explainable}.
In this work, we integrate one of the prominent language models, BERT, as the word-level encoder of our architecture to better represent clinical words.
A closer comparison to our work is FTL-Trans, which implements BERT at the word level and Bi-LSTMs at the note level \cite{zhang2020time}.
To this end, we propose a Hierarchical Transformer Network architecture to encode sequences of clinical notes.
This goes beyond FTL-Trans by both (1) modeling an additional level (more than one document) and (2) utilizing a full stack of Transformers in the model.
We hypothesize this model can learn the contextual complexity of documents, and also leverage structural and temporal information at each level of the hierarchy. 

\section{Model Architecture}
The proposed model architecture is illustrated in Figure~\ref{model}.
The model progressively constructs the representation from the word level towards the final classification level.
The model at each level automatically captures the important parts with multi-headed self attention and accumulates the entire sequence with pooling into the input representation of the next level.
The input length is cropped or padded to a fixed size at the word, sentence, and document levels.
The final representation from the document level is fed to a fully-connected dense layer with a Sigmoid function to output the prediction probability. In the following subsections, we will introduce each model component in detail.
\subsection{Word-level BERT encoders}
\begin{figure}[h]
\centering
\includegraphics[width=\linewidth]{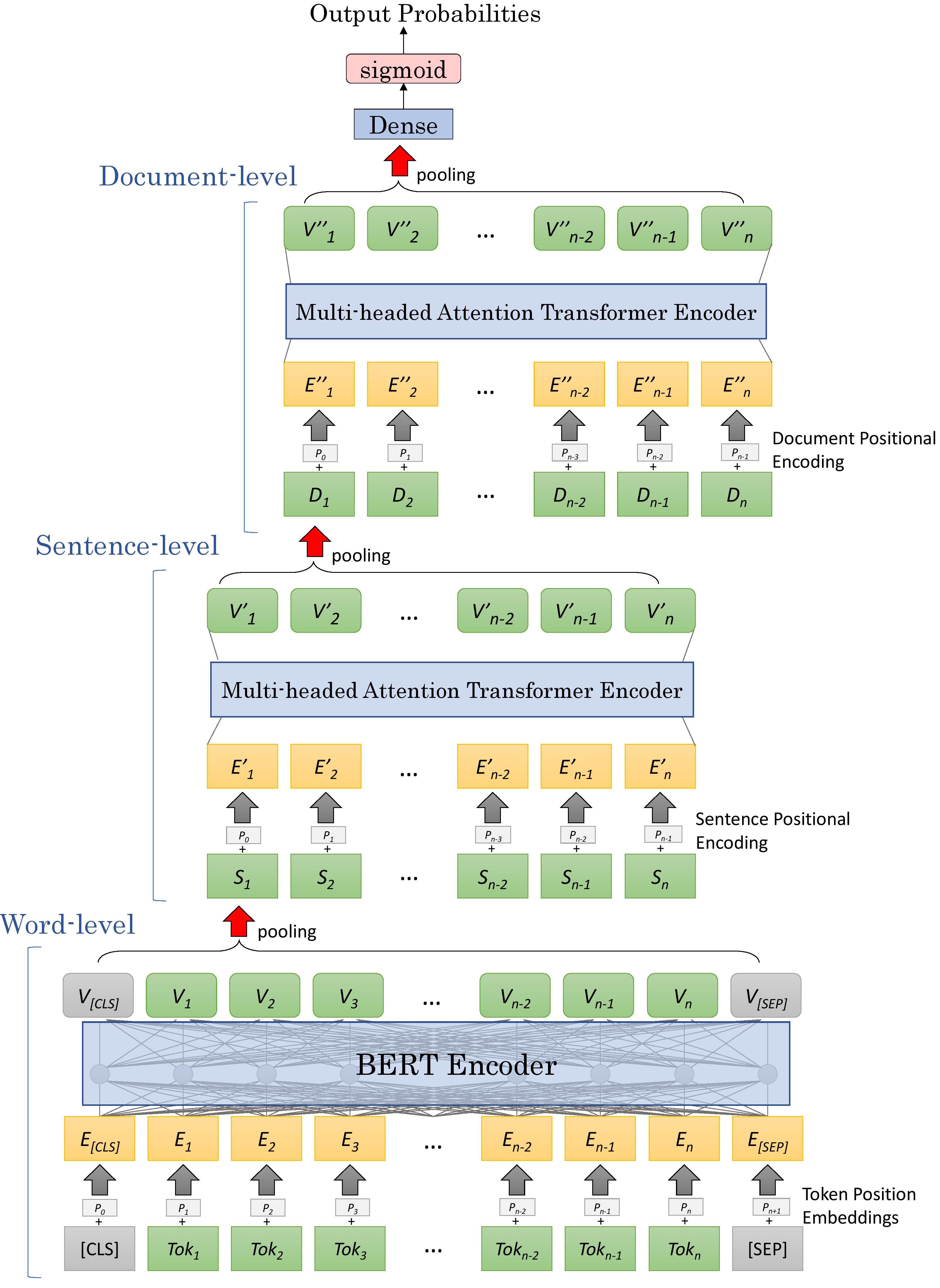}

\caption{Model Architecture}

\label{model}
\end{figure}

As shown in Figure~\ref{model}, at the word level, a BERT model is employed and the word-pieced tokens in a sentence are fed into the model.
We implement the encoder part of the BERT model to represent the words in a sentence, and all parameters in the module are trainable.
Words are preprocessed to obtain the word-pieced tokens through the preprocessing module and with the same token vocabulary list used in BERT \cite{devlin2019bert}. 
Similar to the BERT word-level module, we keep the two special tokens \verb|[CLS]| and \verb|[SEP]| at the start and end of the sentence respectively. 
The first token of each sentence is \verb|[CLS]| and its corresponding hidden state is always considered as the aggregation to represent the entire sentence. 
\verb|[SEP]| is located at the end of the sentence and it is important in differentiating sentences. We omit the segment embeddings and keep the positional encoding. 
Therefore, for a given token $i$, the input embedding $E\textsubscript{i}$ is built by concatenating the word-pieced token embedding \textsl{Tok\textsubscript{i}}, and the positional encoding vector $P\textsubscript{i}$. 

\subsection{Sentence- and Document-level Transformer-based Encoders}\label{pool}
We stack Transformer-based encoders to build the representation of each sentence and each document, respectively.
We briefly introduce the Transformer architecture, and for more details, we recommend the work by \citet{vaswani2017attention}. The Transformer-based encoder is constructed by $N$ layers, and each layer is a residual connection of multi-headed self-attention and a fully-connected feed forward network.
Each self-attention takes three inputs -- $Q$(query), $K$(key), $V$(value) -- to process through the scaled dot-product attention.
The outputs from the scaled dot-product attention are concatenated and put through a linear dense layer.
As opposed to a single self-attention head, $Q$, $K$, and $V$ are partitioned into multiple heads to enable the model to  attend to information at different positions from different representation subviews.

For both sentence- and document-level, positional encoding vectors are concatenated with the input states.
The input state of each sentence is obtained from the first \verb|[CLS]| hidden state of the respective sentence, which is termed as the \textit{CLS-pooling} strategy.
Instead, from sentence to document, and from document to label, we experiment with different pooling strategies for aggregating the representations from previous levels.
This enables providing high levels of the model with more access to the lower-level representations instead of simply using what is accumulated in the \verb|[CLS]| token.
The other pooling strategies we consider consist of \textit{mean}, \textit{max}, and \textit{mean\_max poolings}.
Take the \textit{mean\_max pooling} as an example. The average and maximum of hidden states on the sequence length axis are first obtained separately, and then concatenated to get the pooled output of the whole sequence. 
Following this practice, at the sentence level, the complete sequence of sentences in a given document is pooled to generate the input embeddings of the document. At the document level, the entire series of documents for a given patient is pooled to produce the corresponding patient representation. For the final label, we simply apply a dense layer with a Sigmoid function  to output the classification probabilities. The model is also generalizable to be easily adapted to other machine learning NLP tasks such as pre-training, clustering, and matching, equipped with different loss functions. 
\subsection{Time-aware Adaptive Segmentation and Filling at Document Level}\label{greedy}
Timestamps associated with clinical notes do not always reflect the temporal reality of clinical practice.
Notes often come in bursts and short real-time periods do not inherently have real temporal sequence between each other.
On the other hand, notes outside a long time span contain meaningful sequential information that can be encoded by the neural network.
In order to differentiate short-period co-occurrences with long-range dependencies, we dynamically merge clinical notes into groups to capture the real temporal information between notes.
Meanwhile, such approaches reduce the input sequence length (i.e., number of documents) that are fed into the neural network, which enables the model to learn long-term dependencies more effectively. 
 
For each patient, we first sort the notes in a chronological order, and then apply a greedy algorithm to find the segmentation points.
The algorithm minimizes the maximum time span of contiguous groups. 

Formally, given $T$ documents in a sequence $\left\{d_{t}\right\}_{t=1}^{T}$, we have $k$-$1$ segmentation points $\left\{s_{i}\right\}_{i=1}^{k-1}$ to split the sequence into $k$ groups $\left\{G_{j}\right\}_{j=1}^{k}$, where 
\[
G_{j}=\left \{
\begin{array}{l}
\left\{d_{t}\mid d_{t}.\operatorname{time}<s_{1}\right\}, \hspace{1.1cm} \text{if} \enskip \mathrm{j}=1 
\\ 
\left\{d_{t}\mid d_{t}.\operatorname{time} \geq s_{k-1}\right\}, \hspace{0.7cm} \text{if} \enskip \mathrm{j}=\mathrm{k} 
\\ 
\left\{d_{t} \mid d_{t}.\operatorname{time} \in\left[s_{j-1}, s_{j}\right) \right\}, \text{otherwise}.
\end{array}
\right.
\]
where $d_{t}.\operatorname{time}$ is the charttime  of document $d_{t}$.
\noindent The span of a group is defined as the time difference of the earliest and the latest document in the group:  
\[
\operatorname{span}\left\{G_{j}\right\}=\max_{d_{k} \in G_{j}}\left\{d_{k}.\operatorname{time}\right\}-\min _{d_{k^{\prime}} \in G_{j}}\left\{d_{k^{\prime}}.\operatorname{time}\right\}\]

The optimal choice of the segmentation points can be found by minimizing the following: 
\[
\hat{s}_{1}, \ldots \hat{s}_{k-1}= \argmin_{s_{1}, \ldots s_{k-1}}\{k\} \]
\[\text{subject to} \quad \max_{1 \leq j \leq k}\left\{\operatorname{span}\left(G_{j}\right)\right\} \leq D\] 
\noindent where $D$ constrains the upper bounding of the span. Intuitively, for a given maximum time span, notes within the span are considered as one ``document''.
The notes outside the span are segmented into different units.
In this way, we attempt to preserve the temporal relationship of notes across long terms while combining the notes that come in bursts.
\section{Data and Tasks}
\subsection{Dataset Description}
Our experiments are performed with the MIMIC-III (Medical Information Mart for Intensive Care III) \cite{johnson2016mimic}, which is a de-identified clinical database composed of 46,520 patients with 58,976 admissions in the intensive care units (ICUs). MIMIC-III has been widely studied in clinical NLP tasks as it contains extensive resources of unstructured clinical notes (i.e., 2 million notes in the \verb|NoteEvents| table). We describe note pre-processing in detail in  Appendix~\ref{Note-Preprocessing}.

\subsection{Prediction Tasks}
We evaluate our proposed model to predict in-hospital mortality and phenotypes.
These tasks are standard clinical outcomes of interest that are important to support clinical decisions.
Note that our model is not specifically constrained to these tasks and can be extensively applied to other clinical applications.
Descriptive statistics about patient cohorts are shown in Table~\ref{table1}.

\subsection*{In-hospital Mortality Prediction}

MIMIC-III indicates the time of death for patients who die in the hospital,
enabling us to form the cohorts for in-hospital mortality. 
We use \verb|hospital_expire_flag| (in \verb|Admissions| table) to label positive cases.
In addition, to avoid confusion with multiple admissions of the same patient, we include patients with only one admission.
We exclude \emph{discharge summaries} in mortality prediction because discharge summaries mention the mortality outcome textually.
For the same reason, we also remove all notes with charttime later than the time of death and discharge time.  
\subsection*{Phenotype Prediction}

The purpose of phenotype prediction is to classify patients into a variety of diagnoses.
Specifically, we select the top \textbf{ten} relatively high-prevalence phenotypes, each of which is associated with more than $2000$ patients. 
We consider the diagnostic ICD-9 codes to be the prediction label (a widely-used, though incomplete, surrogate for the phenotype). The phenotype disease name, ICD-9 code, disease type, and the number/percentage of patients for each phenotype in MIMIC-III are reported in Table~\ref{at2}. 
For this task, we include all the notes up to and including the discharge date, because ICD codes are assigned after discharge. 

\begin{table}[t]
\renewcommand{\arraystretch}{1.2}
\caption{Descriptive Statistics of Datasets.}
\label{table1}
\resizebox{0.9\linewidth}{!}{
\begin{tabular}{@{\hskip1pt}l@{\hskip1pt}lcc} \hline
 &  & \begin{tabular}[c]{@{}c@{}}In-hospital \\Mortality\end{tabular} & \begin{tabular}[c]{@{}c@{}}Phenotype \\Prediction\end{tabular}  \\ \hline
\multicolumn{2}{@{\hskip1pt}l}{\begin{tabular}[c]{@{}l@{}}\# Total Patients \\(\% Positives)\end{tabular}}                                              & \begin{tabular}[c]{@{}c@{}}30,881 \\(13.80\%)\end{tabular}     & \begin{tabular}[c]{@{}c@{}}30,990 \\(Table~\ref{at2})\end{tabular}     \\ \hline
\multirow{3}{*}{\begin{tabular}[c]{@{}l@{}}\# Notes \\Per Patient\end{tabular}}  & Mean   & 18.1   & 16.9      \\ 
 & Median  & 12     & 11    \\ 
  & 80 $\%tile$ & 24    & 22    \\ \hline
\multirow{3}{*}{\begin{tabular}[c]{@{}l@{}}\# Sentences \\Per Note\end{tabular}} & Mean     & 29.8    & 37.4         \\ 
  & Median    & 18       & 21        \\ 
   & 80 $\%tile$  & 42   & 50  \\\hline
\multirow{3}{*}{\begin{tabular}[c]{@{}l@{}}\# Wordpieces \\Per Sentence\end{tabular}} & Mean     & 19.2           & 18.9         \\ 
 & Median  & 12              & 12   \\ 
 & 80 $\%tile$  & 22   & 22      \\ \hline
\# Total Sentences  &      & 16,662,894   & 19,656,126  \\ \hline
\multirow{2}{*}{\# Total Notes}    & Raw        & 906,717    & 866,735               \\ & Adaptive  & 559,942  & 525,222    \\\hline
\end{tabular}
}
\begin{tablenotes}
\item \footnotesize \textsuperscript{*} $\%tile$: percentile.
\end{tablenotes}
\end{table}

\section{Experiments}
Here, we describe the compromises made in order to feasibly train such a large model on GPUs, as well as the baselines and evaluation metrics used in the experiments.
Notably, Hierarchical Transformer Networks require smaller BERT models than what are normally used, even when utilizing multiple GPU architectures. To achieve a fast and effective optimization, we implement an exponential decay with linear warmup for learning rate decay.

\subsection{Distributed Training}
The sequence lengths required by our model are significantly longer (many thousands of words) than the standard GPU training can handle without significant compromises (i.e., the standard BERT model has a maximum input length of 512 word pieces).
To resolve resource limits and augment text lengths, we implement the mirrored distribution strategy to distribute the training across multiple GPUs. We introduce the strategy with more details in Appendix~\ref{Mirrored}.
Specifically, we train our proposed model on 4 NVIDIA Tesla V100 GPUs (32G), which means the batch size is quadrupled.
Each training step takes approximately the same time between using $1$ GPU verses using $N\texttt{+}$ GPUs, so the overall time is decreased four-fold if the training takes the same steps. 

\subsection{Compared Baselines}
We compare the proposed model with the following alternative models:
\paragraph{\textsc{BigBird}:} \citet{zaheer2020big} extended the BERT model to longer sequences with sparse attention mechanisms, which is assumed to be the current state-of-the-art method for long-sequence text classification. \textsc{BigBird} achieves good performances with more efficient architectures than traditional BERT, and it uses efficient attention to reduce the complexity while still preserving the model capacity. It can handle sequence lengths up to eight times longer than what was previously possible using similar hardware. In our case, we implement \textsc{BigBird} for each document at the word-level and apply a fully-connected layer for the output probability. Thus, the \textsc{BigBird} utilizes a flattened representation of texts directly from word to label, not considering hierarchical structure. Although it is not a hierarchical model, it can feed the similar input length as the hierarchical models, so we consider this also valuable to be implemented as one baseline.

\paragraph{\textsc{Han}:} The Hierarchical Attention Network (HAN) model is widely used for document classification. We follow \citet{si2020patient} to build the architecture into a triplet structure that encodes notes over a long time (shown in Appendix~\ref{HAN Architecture}).
The model learns the representations at each level with Bi-LSTMs and global context-based attention. 

\paragraph{\textsc{BertLstm}:} We also develop a variation of the proposed model, termed \textsc{BertLstm}, where the Transformers at the sentence and document levels are replaced with Bi-LSTMs. The architecture and model summary is shown in Appendix~\ref{BERT-LSTM Architecture}.
This allows us to measure the absolute performance improvement provided by the top-to-bottom Transformer architecture by replacing the top two Transformer levels with Bi-LSTMs layers. This model is also FTL-Trans \cite{zhang2020time} extended to multiple documents.

To ensure a fair comparison, we enable the hierarchical models (i.e.,\textsc{Han}, \textsc{BertLstm}, and the proposed model) contain the same number of parameters (around 5.6-million), while the \textsc{BigBird} remains the same as in  the released version (because the model is fixed). We carefully select the hyper-parameters to meet this comparison requirement. The detailed descriptions of the model hyper-parameters are described in Appendix~\ref{appendix:hyperparameters}. 
\subsection{Evaluation Metrics}
For method comparisons, we use the Area Under the Receiver Operating Characteristic curve (AUC), the Area Under Precision-Recall curve (PRC), Precision, Recall, and F1-score to report the predictive performance.
The use of PRC in addition to AUC attempts to mitigate variance due to imbalanced class distributions, as the Precision-Recall curve is particularly tailored for identifying less-frequent cases. 
Each cohort is split into train, validation, and test, with a ratio of 8:1:1. We train the model on the train set, apply early stopping on the validation set to prevent overfitting, and report the metrics on the test set.
More specifically, we calculate the loss on the validation set at the end of each epoch (a complete pass over the training data), and early stopping is triggered when the loss has been increasing for three subsequent epochs.


\section{Performance Comparisons}
Table~\ref{comparison} reports the performance comparisons of in-hospital mortality and phenotypes. We observe that our proposed model, Hierarchical Transformer Networks, outperforms other baselines for all tasks in AUC, PRC and F1-score.

We notice the performances of this flattened model, \textsc{Bigbird}, performs considerably worse than the other three hierarchical models in all tasks. So we think a more appropriate use case of \textsc{Bigbird} would be using it for efficient and effective training in long-text document classification. In our case, we have a strong hierarchical structure due to the large number of notes in MIMIC, so the contributions from the hierarchical levels are important.

The performances of \textsc{Han} and \textsc{BertLstm} are approximately the same. The advantages of Hierarchical Transformer Networks over \textsc{BertLstm} are significant in phenotype predictions with improvements of 0.0258 in AUC, 0.0541 in PRC, and 0.0542 in F1-score. And Hierarchical Transformer Networks have relatively small improvements of 0.0251 in AUC, 0.0416 in PRC, and 0.0429 in F1-score, compared to \textsc{Han}. This demonstrates that the Transformers applied at hierarchical levels make a steady contribution to the performance improvement. More importantly, the direct usage of BERT models at the word level has a decisive impact on the predictive performance. Note that we only adopt one layer of encoder in our proposed model, which already yields the best performance across alternatives. According to findings from the  Ablation Study Section~\ref{ablation_study_section}, the model still has room to improve by enlarging the model and incorporating more data. Thus, we believe the great potential of the Hierarchical Transformer Networks would outperform strong state-of-the-art methods in clinical outcome predictions.

We also note that Hierarchical Transformer Networks generate the highest PRCs in in-hospital mortality and almost all phenotype predictions (Table~\ref{auc} b). 
Considering the fact that PRC is a critical metric in clinical problems where properly classifying the positives is important, which is always the case in clinical outcome predictions.
Higher PRC indicates that Hierarchical Transformer Network is more likely to find all the positive cases without accidentally marking negative cases as positive, and such performance is more preferred, especially in clinical phenotype predictions. 
\begin{table}[t]
\centering
\renewcommand{\arraystretch}{1.5}
\caption{Performance comparisons in in-hospital mortality and phenotype predictions. Per-phenotype metrics are shown in Table~\ref{auc}.}
\label{comparison}
\resizebox{1.05\linewidth}{!}{
\begin{tabular}{l|ccccc} 
\hline
& \multicolumn{5}{c}{Macro-AVG of 10-phenotype prediction}           \\
         & AUC    & PRC    & Precision & Recall & F1             \\ 
\hline \hline
\textsc{BigBird}     & 0.7497    &   0.4647  & 0.6513 &   0.3515  & 0.4421               \\
\textsc{Han}   & 0.8845     & 0.6608   &  \textbf{0.7037} &  0.5546 & 0.6033            \\
\textsc{BertLstm}      & 0.8838    & 0.6483   &  0.6712  &  0.5733   & 0.5919      \\
Our Model   & \textbf{0.9096} &  \textbf{0.7024}   &  0.7003  &  \textbf{0.6342}   & \textbf{0.6462}         \\

\hline
\multicolumn{1}{l|}{} & \multicolumn{5}{c}{In-hospital mortality prediction}  \\
                      & AUC    & PRC    & Precision & Recall & F1             \\ 
\hline
\textsc{BigBird}              & 0.8769 & 0.8139 &   0.6924 & 	0.7049 & 	0.6986   \\
\textsc{Han}                   & 0.9610 & 0.8992 &       0.7837 &	\textbf{0.8356} &	0.8088             \\
\textsc{BertLstm}            & 0.9608 & 0.8946 &          0.8740 &	0.7283 &	0.7945          \\
Our Model             & \textbf{0.9677} & \textbf{0.9032} &         \textbf{0.8810} &	0.7603 &	\textbf{0.8162}               \\ 
\hline
\end{tabular}
}

\begin{tablenotes}
\item \small \textsuperscript{*}All models have the same input lengths. \textsc{BertLstm} and Our Model use the same BERT\textsubscript{\textit{\fontfamily{ccr}\selectfont tiny}} at word level.
\end{tablenotes}

\end{table}

\section{Ablation Study} \label{ablation_study_section}
Considerable factor of the Transformer’s success relies on the right setting of hyper-parameters. 
We examine some of the important parameters that impact training performance, robustness, and efficiency to identify an optimal trade-off.
This is critically necessary for our model as the hierarchical transformers require carefully-selected compromises to keep the model size manageable.

\subsection{Input Text Lengths}
The off-the-shelf BERT models are pre-trained with an input sequence length of 128, which is much longer than most sentences in clinical notes. 
As shown in Table~\ref{table1}, the number of word pieces per sentence has a mean value of around 19 (19.2 for the in-hospital mortality cohort, and 18.9 for the phenotype cohort) and a median value of 12.
Thus, it might be a waste of resources to use 128 tokens at the word level.
However, cutting off too many tokens would also harm the pre-trained model capability.
Thus, it would be interesting to evaluate such a trade-off.
We evaluate the performances of \textbf{hypertension} phenotype prediction with varied input sequence lengths at different levels. The results are shown in Table~\ref{seq_len}.

We first examine the results of different sequence lengths at the sentence level, or the number of tokens in a sentence, shown in the last row in Table~\ref{seq_len}.
Even though the sequence length with 128 tokens has reached to 98.6\textsuperscript{th} percentile, the performance does not sizably improve (i.e., from 64 to 128, the AUC slightly increases by 0.0022). 
However, starting from 32, the performances \emph{drop} steadily.
For lengths of 32 and 22, they do not perform well (with AUCs of 0.85 and 0.83) although they reach the 90\textsuperscript{th} and 80\textsuperscript{th} percentiles, respectively.
Thus, we assume that chopping off a large number of tokens out of the original 128 token input, indeed harms the pre-trained model capability.

The results with sequence lengths at the patient and document levels (i.e., the number of notes and sentences) are shown in the Patient and Document columns.
We experiment with 90\textsuperscript{th}, 80\textsuperscript{th}, and 70\textsuperscript{th} percentile data.
All three settings yield an approximately comparable performance with AUC scores around 0.86 to 0.87. 
It is reasonable to have low performance with 70\textsuperscript{th} percentile data (0.86+), but it makes a rather minor difference between 80\textsuperscript{th} and 90\textsuperscript{th} percentiles (0.87+).

\begin{table}[t]
\renewcommand{\arraystretch}{1.3}
\centering
\caption{Performance of hypertension with different input lengths. We denote the first non-header row as the \textbf{base} input, where the models contain 80\textsuperscript{th} percentile data length at the patient and document level, and 64 word pieces at the sentence level.}
\label{seq_len}

\resizebox{\linewidth}{!}{
\begin{tabular}{@{\hskip1pt}l@{\hskip8pt}l@{\hskip8pt}l@{\hskip1pt}|ll}
\hline
\multicolumn{3}{r|}{\renewcommand{\arraystretch}{1}\begin{tabular}[c]{@{}l@{}}Sequence length at each level \\ {\scriptsize[Percentile]}\end{tabular}} & \multicolumn{2}{c}{Hypertension} \\ \hline
Patient & Document & Sentence & \multicolumn{1}{c}{AUC} & \multicolumn{1}{c}{PRC} \\ \hline
22 {\scriptsize[80\textsuperscript{th}]}
& 50  {\scriptsize[80\textsuperscript{th}]} & 
64  {\scriptsize[96.7\textsuperscript{th}]} & 0.8722 & 0.8327 \\ 
 \hline
34  {\scriptsize[90\textsuperscript{th}]} &  &  & 0.8720 & 0.8337 \\
16  {\scriptsize[70\textsuperscript{th}]} &  &  & $\downarrow 0.8623 $ & $\downarrow0.8183$ \\ 
\hline
 & 85  {\scriptsize[90\textsuperscript{th}]} &  & 0.8733 & 0.8299 \\
 & 37  {\scriptsize[70\textsuperscript{th}]} &  & $\downarrow0.8655$ & $\downarrow0.8209$ \\ 
 \hline
  &  & 128  {\scriptsize[98.6\textsuperscript{th}]} & 0.8744 & 0.8309 \\
 &  & 32  {\scriptsize[90\textsuperscript{th}]} & $ \downarrow 0.8546$ & $ \downarrow 0.8147$ \\
 &  & 22  {\scriptsize[80\textsuperscript{th}]} & $\downarrow \downarrow 0.8347 $ & $\downarrow \downarrow0.7997$ \\ 
 \hline
\end{tabular}
}
\begin{tablenotes}
\item \small \textsuperscript{*}Unlisted values are identical to those of the \textbf{base} input.
\end{tablenotes}
\end{table}

\subsection{BERT Variations}

We investigate different distilled BERT models at the word level, including BERT\textsubscript{\textit{\fontfamily{ccr}\selectfont tiny}}, BERT\textsubscript{\textit{\fontfamily{ccr}\selectfont mini}}, BERT\textsubscript{\textit{\fontfamily{ccr}\selectfont small}}, BERT\textsubscript{\textit{\fontfamily{ccr}\selectfont medium}}, BERT\textsubscript{\textit{\fontfamily{ccr}\selectfont base}} \cite{turc2019well}.
The parameter sizes of the models are shown in Appendix~\ref{Model Sizes} Table~\ref{Millions of parameters}. 
Given the same memory limits, we feed into the maximum sequence length for each distilled model, and we investigate if larger models would yield better performance even with smaller input lengths.
As shown in the column \emph{Max Sequence Length} of Table~\ref{bert_size}, different models have varied max input lengths (max\_seq\_len:$D\_S\_W$) that can be incorporated into 4 GPU memories (128G) at maximum capacity. 

Notably, the max document length for BERT\textsubscript{\textit{\fontfamily{ccr}\selectfont medium}} is only 12, but the performance of BERT\textsubscript{\textit{\fontfamily{ccr}\selectfont medium}} achieves the best AUC (0.8869) and PRC (0.8365) among all other combinations. 
For BERT\textsubscript{\textit{\fontfamily{ccr}\selectfont tiny}},  BERT\textsubscript{\textit{\fontfamily{ccr}\selectfont mini}}, and BERT\textsubscript{\textit{\fontfamily{ccr}\selectfont small}}, even though these three models incorporate many more documents than BERT\textsubscript{\textit{\fontfamily{ccr}\selectfont medium}}, the performances of them are still slightly worse than BERT\textsubscript{\textit{\fontfamily{ccr}\selectfont medium}}. 
Interestingly, BERT\textsubscript{\textit{\fontfamily{ccr}\selectfont base}} performs worse than BERT\textsubscript{\textit{\fontfamily{ccr}\selectfont small}} and BERT\textsubscript{\textit{\fontfamily{ccr}\selectfont medium}}.

\begin{table}[t]
\caption{Performance of hypertension with distilled BERT models. Each BERT model is evaluated with three different settings: 1. The maximum length that the memory can afford (\emph{Max Sequence Length}); 2. As BERT\textsubscript{\textit{\fontfamily{ccr}\selectfont base}} incorporates only 6 documents, all the other models are fed with the same 6 documents (\emph{Last Six Notes}); 3. Only discharge summary is fed into the model (\emph{Discharge Summary}).
}
\label{bert_size}
\renewcommand{\arraystretch}{1.3}
\resizebox{0.95\linewidth}{!}{
\begin{tabular}{l|l|cc} 
\hline
       &                                     & \multicolumn{2}{c}{Hypertension}  \\
       & Max Sequence Length                 & AUC             & PRC       \\ 
\hline
BERT\textsubscript{\textit{\fontfamily{ccr}\selectfont tiny}}   & \multicolumn{1}{l|}{D50\_S75\_W128} & 0.8750          &       0.8181         \\
BERT\textsubscript{\textit{\fontfamily{ccr}\selectfont mini}}   & \multicolumn{1}{l|}{D40\_S60\_W64}  & 0.8706          &       0.8066         \\
BERT\textsubscript{\textit{\fontfamily{ccr}\selectfont small}}  & \multicolumn{1}{l|}{D25\_S50\_W64}  & \underline{0.8863}          &        \underline{0.8333}      \\
BERT\textsubscript{\textit{\fontfamily{ccr}\selectfont medium}} & \multicolumn{1}{l|}{D12\_S50\_W64}  & \textcolor{red}{\textbf{0.8869}} &        \textcolor{red}{\textbf{0.8365}}      \\
BERT\textsubscript{\textit{\fontfamily{ccr}\selectfont base}}   & \multicolumn{1}{l|}{D6\_S50\_W64}   & 0.8788          &      0.8178          \\ 
\hline
       & Last Six Notes                      &                 &                \\ 
\hline
BERT\textsubscript{\textit{\fontfamily{ccr}\selectfont tiny}}   &        \multirow{5}{*}{D6\_S50\_W64}                          & 0.8660          &      0.8115          \\
BERT\textsubscript{\textit{\fontfamily{ccr}\selectfont mini}}    &                                      & \underline{0.8776}          &         \underline{0.8213}       \\
BERT\textsubscript{\textit{\fontfamily{ccr}\selectfont small}}   &                                      & 0.8645          &      0.8040          \\
BERT\textsubscript{\textit{\fontfamily{ccr}\selectfont medium}}    &                                   & 0.8763          &         \textbf{0.8231}      \\
BERT\textsubscript{\textit{\fontfamily{ccr}\selectfont base}}  &                                       & \textbf{0.8788}          &       0.8178         \\ 
\hline
       & Discharge Summary                   &                 &                \\ 
\hline
BERT\textsubscript{\textit{\fontfamily{ccr}\selectfont tiny}}  &                \multirow{5}{*}{D1\_S50\_W64}                      & 0.8497          &      0.8030          \\
BERT\textsubscript{\textit{\fontfamily{ccr}\selectfont mini}}   &                                      & 0.8496          &     0.7978          \\
BERT\textsubscript{\textit{\fontfamily{ccr}\selectfont small}}  &                                      & \underline{0.8627}          &      \underline{0.8094}          \\
BERT\textsubscript{\textit{\fontfamily{ccr}\selectfont medium}} &                                     & 0.8503          &       0.8036         \\
BERT\textsubscript{\textit{\fontfamily{ccr}\selectfont base}}   &                                      & \textbf{0.8649}         &        \textbf{0.8161}        \\
\hline
\end{tabular}
}

\begin{tablenotes}
\item \small \textsuperscript{*}All other hyper-parameters are the same across all BERT models. Only the BERT models applied at the word level and the input sequence lengths are different.
\end{tablenotes}
\end{table}

Meanwhile, we investigate the impact of keeping the document length fixed at the BERT\textsubscript{\textit{\fontfamily{ccr}\selectfont base}} max capacity of 6 documents.
We run all other distilled models on the same 6 documents to evaluate if larger models would outperform smaller models given the same amount of input data. 
As presented in the column \emph{Last Six Notes}, we notice that BERT\textsubscript{\textit{\fontfamily{ccr}\selectfont base}} achieves the best AUC and BERT\textsubscript{\textit{\fontfamily{ccr}\selectfont medium}} achieves the best PRC. 

Furthermore, we evaluate our model capacity using only one document to predict the phenotype. We only process the discharge summary to predict whether the patient has hypertension. This would be more challenging than using all the notes because we have a small portion of data. We want to see if the proposed hierarchical architecture can still be used with the same architecture and achieve good performance. As reported in the \emph{Discharge Summary} column, the models continue to perform reasonably well with AUC around 0.85. The best AUC (0.8649) and PRC (0.8161) are achieved by BERT\textsubscript{\textit{\fontfamily{ccr}\selectfont base}}.

However, compared to the performances that extensively use the majority of notes to make predictions, the results using only one note are worse. For all BERT models, the performances with the max sequence length and the last six notes outperform those only using discharge summary.  
Thus, we show the necessity of incorporating as many documents as possible. This is more important when the phenotype is hard to get a satisfactory performance. Adopting all possible notes into the model would yield sufficient room for improvement.

Given the results of the above experiments, along with the general mantra ``more data
and larger models'', we conclude that sufficient data is more crucial and would further improve the performance even if the model size may not be the largest. We therefore provide an applicable recommendation for those cases with less GPU memory: we should first make sure to incorporate sufficient data, then choose the larger model. 

\begin{table}[t]
\centering
\caption{Performance of hypertension predictions: \\(A) numbers of encoder layers, (B) pooling, (C) positional encoding, and (D) adaptive segmentation.}

\label{ablation}
\renewcommand{\arraystretch}{1.3}
\resizebox{0.8\linewidth}{!}{
\begin{tabular}{ll|cc} 
\hline
    &        & \multicolumn{2}{c}{Hypertension}                                                \\ 
\cline{1-4}
    & L                   & AUC    & PRC     \\ 
\cline{1-4}
    & 1                   & 0.8674 & 0.8218     \\
(A) & 2                   & \textbf{0.8722} &\textbf{0.8327}                                              \\
    & 4                   & 0.8645                     & 0.8199                                             \\
    & 6                   & 0.8672                     & 0.8213                                              \\
    & 8                   & 0.8684                     & 0.8285                                             \\ 
\hline
    & pooling             &                            &                                                   \\ 
\cline{2-2}
    & first               & 0.8683                     & 0.8214                                              \\
(B) & mean                & 0.8702                     & 0.8295                                             \\
    & max                 & 0.8675                     & 0.8222                                              \\
    & mean\_max           & \textbf{0.8722} &\textbf{0.8327}                                            \\ 
\hline
\multirow{2}{*}{(C)} & w/o                 & 0.8700                     & 0.8294                                            \\
    & positional encoding & \textbf{0.8722} &\textbf{0.8327}                                             \\ 
\hline
\multirow{2}{*}{(D)} & w/o                 & 0.8558                     & 0.7887                                              \\
    & adaptive segment    & \textbf{0.8722} &\textbf{0.8327}                                              \\
\hline
\end{tabular}
}
\begin{tablenotes}
\item {\small \textsuperscript{*}Unless specified, other hyper-parameters identical to best-performing model.\par}
\end{tablenotes}
\end{table}

\subsection{Transformer Encoder Variations}

We first evaluate the performance with different \textbf{numbers of encoder layers}
($L$ = $1,2,4,6,8$) in the sentence- and document-level transformers.
Table~\ref{ablation}(A) shows that the model with 2 encoder layers achieves the best AUC (0.8722) and PRC (0.8327). 
Notably, models with fewer layers ($L$=$1,2$) generally perform better than those with more layers ($L$=$4,6$).
Although this is opposed to the general mantra that larger models yield better performance, we assume it is because extreme model sizes might lead to an improvement bottleneck if the model is only used as fine-tuning classification.

We also compare different \textbf{pooling strategies} of how to aggregate the representations from the previous to the next level.
Table~\ref{ablation}(B) finds that mean\_max pooling is the best-performing pooling method.

As shown in Table~\ref{ablation}(C), excluding \textbf{positional encodings} slightly hurts performance.
Thus, position-sensitive information is necessary for each representation unit to incorporate the orders of words/sentences/documents.

The results in Table~\ref{ablation}(D) show that there are significant decreases in AUC and PRC if we remove the \textbf{adaptive segmentation}. If clinical notes for the same patient are all independent without proper segmentation, the effect is clearly reflected in the performance (0.8558 in AUC and 0.7887 in PRC).

\section{Conclusion}
In this work, we develop the Hierarchical Transformer Network to effectively process the sequential and hierarchical structure of clinical notes. The model takes the interrelations among clinical notes and the multilevel hierarchical information into account. We evaluate our approach using common clinical predictions, including in-hospital mortality and phenotype predictions. Our results demonstrate that the proposed model outperforms strong baselines in AUC, PRC and F1-score for both predictions. We also perform an extensive range of experiments on the proposed model with an optimal trade-off to achieve robust and effective training given computational resource limits.  

\newpage

\bibliography{anthology,custom}

\begin{thebibliography}{33}
\expandafter\ifx\csname natexlab\endcsname\relax\def\natexlab#1{#1}\fi

\bibitem[{Adhikari et~al.(2019)Adhikari, Ram, Tang, and
  Lin}]{adhikari_docbert_2019}
Ashutosh Adhikari, Achyudh Ram, Raphael Tang, and Jimmy Lin. 2019.
\newblock {DocBERT: BERT for document classification}.
\newblock \emph{arXiv}, 1904.08398.

\bibitem[{Alsentzer et~al.(2019)Alsentzer, Murphy, Boag, Weng, Jindi, Naumann,
  and McDermott}]{alsentzer2019publicly}
Emily Alsentzer, John Murphy, William Boag, Wei-Hung Weng, Di~Jindi, Tristan
  Naumann, and Matthew McDermott. 2019.
\newblock {Publicly Available Clinical BERT Embeddings}.
\newblock In \emph{Proceedings of the 2nd Clinical Natural Language Processing
  Workshop}, pages 72--78.

\bibitem[{Beltagy et~al.(2019)Beltagy, Lo, and Cohan}]{beltagy2019scibert}
Iz~Beltagy, Kyle Lo, and Arman Cohan. 2019.
\newblock {SciBERT: A Pretrained Language Model for Scientific Text}.
\newblock In \emph{Proceedings of the 2019 Conference on Empirical Methods in
  Natural Language Processing and the 9th International Joint Conference on
  Natural Language Processing (EMNLP-IJCNLP)}, pages 3606--3611.

\bibitem[{Choi et~al.(2020)Choi, Xu, Li, Dusenberry, Flores, Xue, and
  Dai}]{choi2020learning}
Edward Choi, Zhen Xu, Yujia Li, Michael Dusenberry, Gerardo Flores, Emily Xue,
  and Andrew Dai. 2020.
\newblock {Learning the graphical structure of electronic health records with
  graph convolutional transformer}.
\newblock In \emph{Proceedings of the AAAI Conference on Artificial
  Intelligence}, volume~34, pages 606--613.

\bibitem[{Devlin et~al.(2019)Devlin, Chang, Lee, and
  Toutanova}]{devlin2019bert}
Jacob Devlin, Ming-Wei Chang, Kenton Lee, and Kristina Toutanova. 2019.
\newblock {BERT: Pre-training of Deep Bidirectional Transformers for Language
  Understanding}.
\newblock In \emph{Proceedings of the 2019 Conference of the North American
  Chapter of the Association for Computational Linguistics: Human Language
  Technologies, Volume 1 (Long and Short Papers)}, pages 4171--4186.

\bibitem[{Feng et~al.(2020)Feng, Shaib, and
  Rudzicz}]{feng-etal-2020-explainable}
Jinyue Feng, Chantal Shaib, and Frank Rudzicz. 2020.
\newblock Explainable clinical decision support from text.
\newblock In \emph{Proceedings of the 2020 Conference on Empirical Methods in
  Natural Language Processing (EMNLP)}, pages 1478--1489, Online. Association
  for Computational Linguistics.

\bibitem[{Gao et~al.(2018)Gao, Ramanathan, and Tourassi}]{gao2018hierarchical}
Shang Gao, Arvind Ramanathan, and Georgia Tourassi. 2018.
\newblock {Hierarchical convolutional attention networks for text
  classification}.
\newblock In \emph{Proceedings of The Third Workshop on Representation Learning
  for NLP}, pages 11--23.

\bibitem[{Johnson et~al.(2016)Johnson, Pollard, Shen, Li-Wei, Feng, Ghassemi,
  Moody, Szolovits, Celi, and Mark}]{johnson2016mimic}
Alistair~EW Johnson, Tom~J Pollard, Lu~Shen, H~Lehman Li-Wei, Mengling Feng,
  Mohammad Ghassemi, Benjamin Moody, Peter Szolovits, Leo~Anthony Celi, and
  Roger~G Mark. 2016.
\newblock {MIMIC-III, a freely accessible critical care database}.
\newblock \emph{Scientific data}, 3(1):1--9.

\bibitem[{Joshi et~al.(2020)Joshi, Chen, Liu, Weld, Zettlemoyer, and
  Levy}]{joshi2020spanbert}
Mandar Joshi, Danqi Chen, Yinhan Liu, Daniel~S Weld, Luke Zettlemoyer, and Omer
  Levy. 2020.
\newblock {SpanBERT: Improving pre-training by representing and predicting
  spans}.
\newblock \emph{Transactions of the Association for Computational Linguistics},
  8:64--77.

\bibitem[{Kowsari et~al.(2017)Kowsari, Brown, Heidarysafa, Meimandi, Gerber,
  and Barnes}]{kowsari2017hdltex}
Kamran Kowsari, Donald~E Brown, Mojtaba Heidarysafa, Kiana~Jafari Meimandi,
  Matthew~S Gerber, and Laura~E Barnes. 2017.
\newblock {HDLTex: Hierarchical deep learning for text classification}.
\newblock In \emph{2017 16th IEEE international conference on machine learning
  and applications (ICMLA)}, pages 364--371. IEEE.

\bibitem[{Lan et~al.(2019)Lan, Chen, Goodman, Gimpel, Sharma, and
  Soricut}]{lan2019albert}
Zhenzhong Lan, Mingda Chen, Sebastian Goodman, Kevin Gimpel, Piyush Sharma, and
  Radu Soricut. 2019.
\newblock {AlBERT: A lite bert for self-supervised learning of language
  representations}.
\newblock \emph{arXiv preprint arXiv:1909.11942}.

\bibitem[{Lee et~al.(2020)Lee, Yoon, Kim, Kim, Kim, So, and
  Kang}]{lee2020biobert}
Jinhyuk Lee, Wonjin Yoon, Sungdong Kim, Donghyeon Kim, Sunkyu Kim, Chan~Ho So,
  and Jaewoo Kang. 2020.
\newblock {BioBERT: a pre-trained biomedical language representation model for
  biomedical text mining}.
\newblock \emph{Bioinformatics}, 36(4):1234--1240.

\bibitem[{Li and Yu(2020)}]{Li_Yu_2020}
Fei Li and Hong Yu. 2020.
\newblock {ICD Coding from Clinical Text Using Multi-Filter Residual
  Convolutional Neural Network}.
\newblock \emph{Proceedings of the AAAI Conference on Artificial Intelligence},
  34(05):8180--8187.

\bibitem[{Li et~al.(2020)Li, Rao, Solares, Hassaine, Ramakrishnan, Canoy, Zhu,
  Rahimi, and Salimi-Khorshidi}]{li2020behrt}
Yikuan Li, Shishir Rao, Jos{\'e} Roberto~Ayala Solares, Abdelaali Hassaine,
  Rema Ramakrishnan, Dexter Canoy, Yajie Zhu, Kazem Rahimi, and Gholamreza
  Salimi-Khorshidi. 2020.
\newblock {BEHRT: transformer for electronic health records}.
\newblock \emph{Scientific reports}, 10(1):1--12.

\bibitem[{Liu et~al.(2018)Liu, Zhang, and Razavian}]{pmlr-v85-liu18b}
Jingshu Liu, Zachariah Zhang, and Narges Razavian. 2018.
\newblock Deep ehr: Chronic disease prediction using medical notes.
\newblock In \emph{Proceedings of the 3rd Machine Learning for Healthcare
  Conference}, volume~85 of \emph{Proceedings of Machine Learning Research},
  pages 440--464, Palo Alto, California. PMLR.

\bibitem[{Liu et~al.(2019)Liu, Ott, Goyal, Du, Joshi, Chen, Levy, Lewis,
  Zettlemoyer, and Stoyanov}]{liu2019roberta}
Yinhan Liu, Myle Ott, Naman Goyal, Jingfei Du, Mandar Joshi, Danqi Chen, Omer
  Levy, Mike Lewis, Luke Zettlemoyer, and Veselin Stoyanov. 2019.
\newblock {RoBERTa: A robustly optimized bert pretraining approach}.
\newblock \emph{arXiv preprint arXiv:1907.11692}.

\bibitem[{Makarenkov and Rokach(2020)}]{makarenkov2020lessons}
Victor Makarenkov and Lior Rokach. 2020.
\newblock {Lessons Learned from Applying off-the-shelf BERT: There is no
  SilverBullet}.
\newblock \emph{arXiv preprint arXiv:2009.07238}.

\bibitem[{Pappagari et~al.(2019)Pappagari, Zelasko, Villalba, Carmiel, and
  Dehak}]{pappagari_hierarchical_2019}
Raghavendra Pappagari, Piotr Zelasko, Jesús Villalba, Yishay Carmiel, and
  Najim Dehak. 2019.
\newblock {Hierarchical transformers for long document classification}.
\newblock In \emph{2019 {IEEE} Automatic Speech Recognition and Understanding
  Workshop ({ASRU})}, pages 838--844. {IEEE}.

\bibitem[{Peng et~al.(2019)Peng, Yan, and Lu}]{peng2019transfer}
Yifan Peng, Shankai Yan, and Zhiyong Lu. 2019.
\newblock {Transfer Learning in Biomedical Natural Language Processing: An
  Evaluation of BERT and ELMo on Ten Benchmarking Datasets}.
\newblock In \emph{Proceedings of the 18th BioNLP Workshop and Shared Task},
  pages 58--65.

\bibitem[{Popel and Bojar(2018)}]{popel_training_2018}
Martin Popel and Ondřej Bojar. 2018.
\newblock {Training tips for the transformer model}.
\newblock 110(1):43--70.
\newblock Publisher: Sciendo.

\bibitem[{Sanh et~al.(2019)Sanh, Debut, Chaumond, and
  Wolf}]{sanh2019distilbert}
Victor Sanh, Lysandre Debut, Julien Chaumond, and Thomas Wolf. 2019.
\newblock {DistilBERT, a distilled version of BERT: smaller, faster, cheaper
  and lighter}.
\newblock \emph{arXiv preprint arXiv:1910.01108}.

\bibitem[{Si et~al.(2021)Si, Du, Li, Jiang, Miller, Wang, {Jim Zheng}, and
  Roberts}]{SI2021103671}
Yuqi Si, Jingcheng Du, Zhao Li, Xiaoqian Jiang, Timothy Miller, Fei Wang,
  W.~{Jim Zheng}, and Kirk Roberts. 2021.
\newblock Deep representation learning of patient data from electronic health
  records (ehr): A systematic review.
\newblock \emph{Journal of Biomedical Informatics}, 115:103671.

\bibitem[{Si and Roberts(2020)}]{si2020patient}
Yuqi Si and Kirk Roberts. 2020.
\newblock {Patient Representation Transfer Learning from Clinical Notes based
  on Hierarchical Attention Network}.
\newblock \emph{AMIA Summits on Translational Science Proceedings}, 2020:597.

\bibitem[{Si et~al.(2019)Si, Wang, Xu, and Roberts}]{si2019enhancing}
Yuqi Si, Jingqi Wang, Hua Xu, and Kirk Roberts. 2019.
\newblock {Enhancing clinical concept extraction with contextual embeddings}.
\newblock \emph{Journal of the American Medical Informatics Association},
  26(11):1297--1304.

\bibitem[{Song et~al.(2018)Song, Rajan, Thiagarajan, and
  Spanias}]{song2018attend}
Huan Song, Deepta Rajan, Jayaraman Thiagarajan, and Andreas Spanias. 2018.
\newblock {Attend and diagnose: Clinical time series analysis using attention
  models}.
\newblock In \emph{Proceedings of the AAAI Conference on Artificial
  Intelligence}, volume~32.

\bibitem[{Turc et~al.(2019)Turc, Chang, Lee, and Toutanova}]{turc2019well}
Iulia Turc, Ming-Wei Chang, Kenton Lee, and Kristina Toutanova. 2019.
\newblock Well-read students learn better: On the importance of pre-training
  compact models.
\newblock \emph{arXiv preprint arXiv:1908.08962}.

\bibitem[{Vaswani et~al.(2017)Vaswani, Shazeer, Parmar, Uszkoreit, Jones,
  Gomez, Kaiser, and Polosukhin}]{vaswani2017attention}
Ashish Vaswani, Noam Shazeer, Niki Parmar, Jakob Uszkoreit, Llion Jones,
  Aidan~N Gomez, {\L}ukasz Kaiser, and Illia Polosukhin. 2017.
\newblock {Attention is all you need}.
\newblock In \emph{Proceedings of the 31st International Conference on Neural
  Information Processing Systems}, pages 6000--6010.

\bibitem[{Yang et~al.(2020)Yang, Zhang, Li, Bendersky, and
  Najork}]{yang_beyond_2020}
Liu Yang, Mingyang Zhang, Cheng Li, Michael Bendersky, and Marc Najork. 2020.
\newblock {Beyond 512 Tokens: Siamese Multi-depth Transformer-based
  Hierarchical Encoder for Long-Form Document Matching}.
\newblock In \emph{Proceedings of the 29th {ACM} International Conference on
  Information \& Knowledge Management}, pages 1725--1734.

\bibitem[{Yang et~al.(2016)Yang, Yang, Dyer, He, Smola, and
  Hovy}]{yang2016hierarchical}
Zichao Yang, Diyi Yang, Chris Dyer, Xiaodong He, Alex Smola, and Eduard Hovy.
  2016.
\newblock {Hierarchical attention networks for document classification}.
\newblock In \emph{Proceedings of the 2016 conference of the North American
  chapter of the association for computational linguistics: human language
  technologies}, pages 1480--1489.

\bibitem[{Zaheer et~al.(2020)Zaheer, Guruganesh, Dubey, Ainslie, Alberti,
  Ontanon, Pham, Ravula, Wang, Yang et~al.}]{zaheer2020big}
Manzil Zaheer, Guru Guruganesh, Kumar~Avinava Dubey, Joshua Ainslie, Chris
  Alberti, Santiago Ontanon, Philip Pham, Anirudh Ravula, Qifan Wang, Li~Yang,
  et~al. 2020.
\newblock Big bird: Transformers for longer sequences.
\newblock In \emph{NeurIPS}.

\bibitem[{Zhang et~al.(2020)Zhang, Thadajarassiri, Sen, and
  Rundensteiner}]{zhang2020time}
Dongyu Zhang, Jidapa Thadajarassiri, Cansu Sen, and Elke Rundensteiner. 2020.
\newblock {Time-Aware Transformer-based Network for Clinical Notes Series
  Prediction}.
\newblock In \emph{Machine Learning for Healthcare Conference}, pages 566--588.
  PMLR.

\bibitem[{Zhang et~al.(2019)Zhang, Wei, and Zhou}]{zhang_hibert_2019}
Xingxing Zhang, Furu Wei, and Ming Zhou. 2019.
\newblock {HIBERT: Document Level Pre-training of Hierarchical Bidirectional
  Transformers for Document Summarization}.
\newblock In \emph{Proceedings of the 57th Annual Meeting of the Association
  for Computational Linguistics}, pages 5059--5069.

\bibitem[{Zhou et~al.(2016)Zhou, Wan, and Xiao}]{zhou2016attention}
Xinjie Zhou, Xiaojun Wan, and Jianguo Xiao. 2016.
\newblock {Attention-based LSTM network for cross-lingual sentiment
  classification}.
\newblock In \emph{Proceedings of the 2016 conference on empirical methods in
  natural language processing}, pages 247--256.

\end{thebibliography}
\bibliographystyle{acl_natbib}

\appendix
\counterwithin{figure}{section}
\counterwithin{table}{section}

\newpage
\section{Appendix}
\subsection{Note Preprocessing} \label{Note-Preprocessing}

For all predictions, we keep patients more than 18 years old. We consider each note entry in \verb|NoteEvents| as a single note.
Notes labeled with \verb|ISERROR| tags and blank entries are excluded.
Notes are sorted in ascending order by \verb|charttime|. 
For each patient, notes are segmented/filled according to Section~\ref{greedy}.
Sentence segmentation is performed simply using periods and newline characters.
(It results in highly sub-optimal sentence segmentation, but this is a very challenging problem on clinical notes.)
Regular expressions are applied to remove special tokens including masked Protected Health Information (PHI) and numerical digits.
Even though such tokens can be matched with BERT word-pieced vocabularies, these special characters would occupy space in sentences and overall provide less meaningful information related to the clinical prediction tasks. 
\subsection{Mirrored Strategy}  \label{Mirrored}

The mirrored distribution strategy is developed with data parallelism, where the same model is replicated on multiple GPU devices on a single machine and different slices of the input data are fed into them accordingly.  
The model variables on each GPU will be mirrored and trained independently in sync.
After each epoch of training, the learned variables are aggregated across each of the GPUs using an all-reduce algorithm by NVIDIA NCCL.
\subsection{Model Hyper-parameter and Architecture}
\label{appendix:hyperparameters}

We introduce the hyper-parameter of each model in the baselines and the proposed model in this section. Note that except \textsc{BigBird}, 
we enable the compared models contain the similar number of parameters to ensure the fairness of the comparison.


\paragraph{Hierarchical Transformer Network:} \label{HTN Architecture}
We denote $L$ as the number of layers in the encoder, $num\_heads$ as the number of parallel heads in multi-headed attention, $d\textsubscript{model}$ as the dimension of hidden units, and $d\textsubscript{ff}$ as the dimensions of the position-wise feed forward networks. 
At the word level, we experiment with a series of smaller uncased BERT models with distilled knowledge including BERT\textsubscript{\textit{\fontfamily{ccr}\selectfont tiny}}, BERT\textsubscript{\textit{\fontfamily{ccr}\selectfont mini}}, BERT\textsubscript{\textit{\fontfamily{ccr}\selectfont small}}, BERT\textsubscript{\textit{\fontfamily{ccr}\selectfont medium}}, BERT\textsubscript{\textit{\fontfamily{ccr}\selectfont base}} \cite{turc2019well}.
The BERT models are downloaded from TensorFlow Hub\footnote{\url{https://tfhub.dev/}} to be used as a trainable component directly.
For instance, BERT\textsubscript{\textit{\fontfamily{ccr}\selectfont tiny}} is a two-layer encoder ($L=2$) with a 2-head self-attention ($num\_heads=2$), and produces an output embedding with a hidden size of 128 ($d\textsubscript{model} = 128$). 
\begin{table}[t]
\centering
\renewcommand{\arraystretch}{0.85}
\caption{Hyper-parameter of the Hierarchial Transformer Networks}
\label{htn_params}
\resizebox{0.7\linewidth}{!}{%
\begin{tabular}{llr}
\hline
 & \textbf{param\_name} & \textbf{value} \\ \hline
\multirow{3}{*}{word\_level} & num\_layers & 2 \\
 & d\_model & 128 \\
 & num\_heads & 2 \\ \hline
\multirow{5}{*}{\begin{tabular}[c]{@{}l@{}}sentence-\\ document-\\ levels\end{tabular}} & num\_layers & 1 \\
 & d\_model & 128 \\
 & num\_heads & 8 \\
 & dff & 2048 \\
 & dropout & 0.2 \\ \hline
\end{tabular}
}
\end{table}

\begin{figure}[t]
\centering
\includegraphics[width=0.8\linewidth]{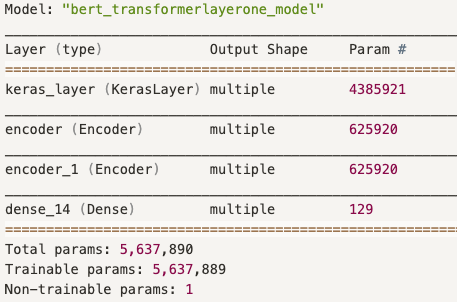}
\caption{Model Summary of the Hierarchical Transformer Network with One Encoder Layer}
\label{keras-berttrans}
\end{figure}
At the sentence and document levels, we keep the encoder with the same hidden unit size as the BERT model.
That is, if BERT\textsubscript{\textit{\fontfamily{ccr}\selectfont tiny}} is used at the word level, $d\textsubscript{model} = 128$ at both the sentence and document levels.
We set the default values from Transformer\textsubscript{\textit{\fontfamily{ccr}\selectfont base}} \cite{vaswani2017attention} for other hyper-parameters as follows: $num\_heads=8$, $d\textsubscript{ff}=2048$, input position encoding dimension is the same with $d\textsubscript{model}$, layer normalization $\varepsilon=1e-6$, and dropout rate $P\textsubscript{drop}=0.2$. The detailed hyper-parameter of the proposed model is shown in Table~\ref{htn_params}.

The models are trained with the Adam optimizer. More importantly, to achieve a fast and effective optimization, we implement an exponential decay with linear warmup for learning rate decay.

For the model that is specifically used in the performance comparison, we adopt an one-layer encoder both at the sentence and document levels, so that the model has around 5.6M parameters. The detailed summary of the proposed model architecture is shown in Figure~\ref{keras-berttrans}.

\begin{table}[h]
\centering
\renewcommand{\arraystretch}{0.8}
\caption{Hyper-parameter of the \textsc{BigBird} model}
\label{bigbird_params}
\resizebox{0.7\linewidth}{!}{%
\begin{tabular}{lr}
\hline
\textbf{param\_name} & \textbf{value} \\ \hline
attention\_probs\_dropout\_prob & 0.1 \\
hidden\_act & gelu \\
hidden\_dropout\_prob & 0.1 \\
hidden\_size & 768 \\
initializer\_range & 0.02 \\
intermediate\_size & 3072 \\
max\_position\_embeddings & 4096 \\
num\_attention\_heads & 12 \\
num\_hidden\_layers & 12 \\
type\_vocab\_size & 2 \\
scope & bigbird \\
use\_bias & TRUE \\
rescale\_embedding & FALSE \\
use\_gradient\_checkpointing & FALSE \\
attention\_type & block\_sparse \\
norm\_type & postnorm \\
block\_size & 16 \\
num\_rand\_blocks & 3 \\
max\_encoder\_length & 1024 \\
vocab\_size & 50358 \\ \hline
\end{tabular}}
\end{table}

\begin{figure}[h]
\centering
\includegraphics[width=0.7\linewidth]{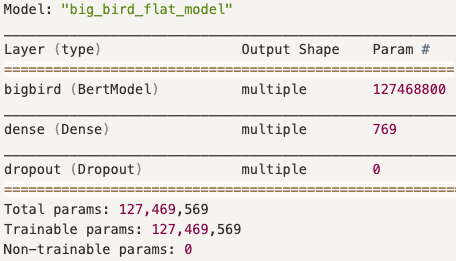}
\caption{Model Summary of the  \textsc{BigBird}}
\label{keras-bigbird}
\end{figure}

\paragraph{\textsc{BigBird}:} \label{BigBird Architecture} 
It is a sparse-attention based transformer model that allows to handle significantly longer sequences than the original BERT model. \textsc{BigBird} also adopts global and random attentions to a more computationally efficient attention mechanism. It shows such attentions closely resemble the full attention in BERT models. \textsc{BigBird} also improve the performance on a wide variety of NLP tasks as a result of its capacity feeding into more input sequences. 
We apply the \textsc{BigBird} for each document at the word-level. In other words, each clinical note is fed into the \textsc{BigBird} from words. The hidden output from \textsc{BigBird} for each note is then fed into a fully-connected network for the final classification. Although this pipeline is not the same with other compared baselines and the proposed model (flattened vs hierarchical), we assume this workflow is the current best way to implement \textsc{BigBird} at patient-level classification (based on our preliminary experiment results). In the future, we will further investigate into how to implement \textsc{BigBird} into a hierarchical manner. 

The detailed hyper-parameter of \textsc{BigBird} is reported in Table~\ref{bigbird_params}. We also implement an exponential decay with linear warmup for the learning rate decay. The detailed model summary is shown in Figure~\ref{keras-bigbird}.

\begin{figure}[t]
\centering
\includegraphics[width=0.85\linewidth]{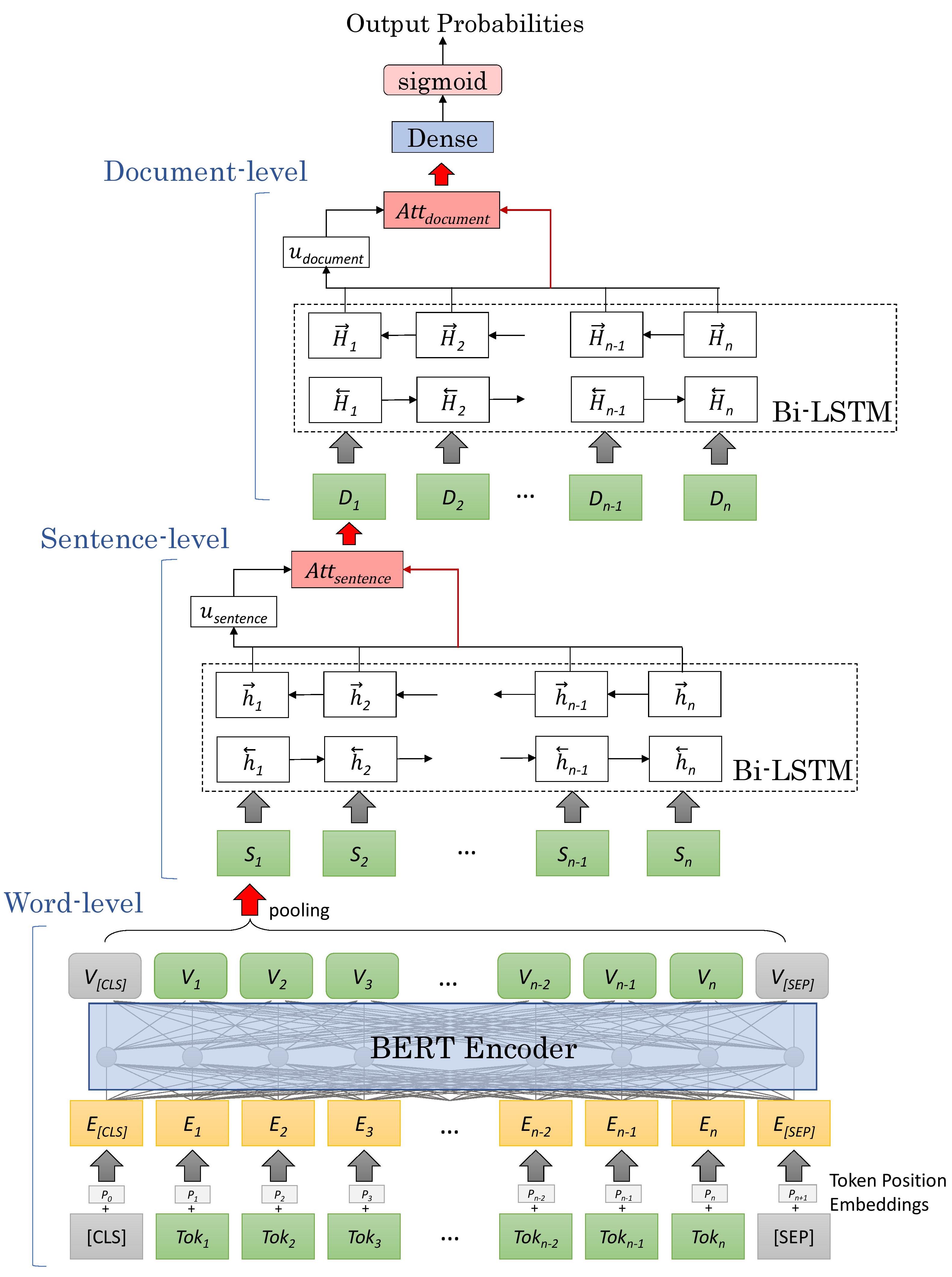}
\caption{The \textsc{BertLstm} Model Architecture}
\label{bert_lstm}
\end{figure}

\begin{figure}[t]
\centering
\includegraphics[width=0.8\linewidth]{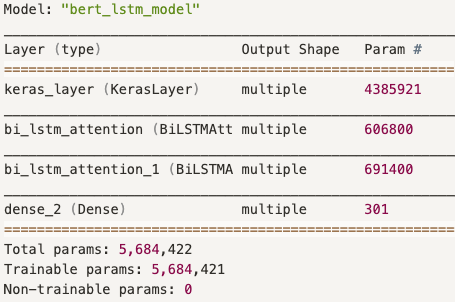}
\caption{Model Summary of the  \textsc{BertLstm}}
\label{keras-bertlstm}
\end{figure}

\paragraph{\textsc{BertLstm}} \label{BERT-LSTM Architecture}
The architecture and model summary of \textsc{BertLstm} is shown in Figure~\ref{bert_lstm} and Figure~\ref{keras-bertlstm}, respectively. The word level still maintains a BERT model as a fully-trainable component. The sentence and document sequential information are encoded through Bi-LSTM. A global context-based attention is also adopted to capture the important knowledge and aggregate the embeddings from the previous level to the next level.

The BERT size in \textsc{BertLstm} is the same with the proposed model at the word level (BERT\textsubscript{\textit{\fontfamily{ccr}\selectfont tiny}}). The Bi-LSTM in \textsc{BertLstm} takes a hidden unit size of 200 and 150 at the sentence and document level, respectively. The output size at the document and patient level is 200 and 100, respectively. 

\begin{figure}[t]
\centering
\includegraphics[width=0.8\linewidth]{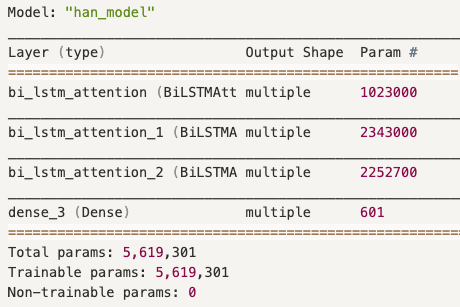}
\caption{Model Summary of the  \textsc{Han}}
\label{keras-han}
\end{figure}

\paragraph{\textsc{Han}:} \label{HAN Architecture} 
The \textsc{Han} is the same with Hierarchical Transformer Network where three layers of networks progressively build from word to sentence, sentence to document, and document to patient. The only difference is that we replace Transformers with Bi-LSTM for the \textsc{Han} model at all layers. For Bi-LSTM in \textsc{Han}, we use a hidden unit size of 300 for all three levels. The output size at the sentence, document, patient level is 300, 300, and 150, respectively. The model summary of \textsc{Han} is shown in Figure~\ref{keras-han}. 

\subsection{Distilled BERT Model Sizes}  \label{Model Sizes}
The model sizes with different word-level BERT models and various numbers of sentence and document transformer layers are in Table~\ref{Millions of parameters}.
\begin{table}[h]
\centering
\caption{Millions of parameters.}
\renewcommand{\arraystretch}{1.1}
\label{Millions of parameters}
\resizebox{\linewidth}{!}{%
\begin{tabular}{l|ccccc}
\hline
 \diagbox{$L^1$}{size} & \begin{tabular}[c]{@{}c@{}}BERT\textsubscript{\textit{\fontfamily{ccr}\selectfont tiny}} \\ (4.4M)\end{tabular} & \begin{tabular}[c]{@{}c@{}}BERT\textsubscript{\textit{\fontfamily{ccr}\selectfont mini}} \\ (11.2M)\end{tabular} & \begin{tabular}[c]{@{}c@{}}BERT\textsubscript{\textit{\fontfamily{ccr}\selectfont small}} \\ (28.8M)\end{tabular} & \begin{tabular}[c]{@{}c@{}}BERT\textsubscript{\textit{\fontfamily{ccr}\selectfont medium}} \\ (41.4M)\end{tabular} & \begin{tabular}[c]{@{}c@{}}BERT\textsubscript{\textit{\fontfamily{ccr}\selectfont base}} \\ (110M)\end{tabular} \\ \hline
1 & 5.6 & 13.9 & 35.6 & 48.2 & 121.7 \\
2 & 6.8 & 16.7 & 42.4 & 55 & 133.9 \\
4 & 9.2 & 22.2 & 56.1 & 68.7 & 158.3 \\
6 & 11.6 & 27.7 & 69.7 & 82.4 & 182.7 \\
8 & 13.9 & 33.3 & 83.4 & 96 & 207.1 \\ \hline
\end{tabular}}
\begin{tablenotes}
\item \small \textsuperscript{1}$L$: number of encoder layers at sentence and document level. 
\end{tablenotes}
\end{table}

\subsection{Parameter Allocation Experiments}
We explore the effect of allocating memory to different levels of the hierarchy. 
to assess impact on performance.
That is, given the same memory constraints and parameter sizes, we examine which level of the Hierarchical Transformer Network should be provided with more resources: the upper levels in documents and sentences, or the lower word level; and whether such allocation would impact the performance.

We train BERT\textsubscript{\textit{\fontfamily{ccr}\selectfont tiny}}\_L$8$ and BERT\textsubscript{\textit{\fontfamily{ccr}\selectfont mini}}\_L$1$, both of which have 13.9-million parameters. 
BERT\textsubscript{\textit{\fontfamily{ccr}\selectfont tiny}}\_L$8$ allocates more to the document and sentence levels with deep encoders ($L$=$8$), but has only two layers of encoder at the word level (built in BERT\textsubscript{\textit{\fontfamily{ccr}\selectfont tiny}}).
While BERT\textsubscript{\textit{\fontfamily{ccr}\selectfont mini}}\_L$1$ allocates more to the word level with 4 layers (built in BERT\textsubscript{\textit{\fontfamily{ccr}\selectfont mini}}), but has only one layer of encoders at document and sentence levels.

Table~\ref{tinybertL8_minibertL1} shows training with
deeper layers at the word level achieves slightly better performance than deeper layers at upper levels with the same overall model size. It indicates the hierarchical model reaches good results by focusing largely on the word layer and capturing the underlying low-level features in language, at least for phenotype classifications (perhaps other tasks may require more emphasis on higher-level representation).

\begin{table}[h]

\caption{Allocation at different hierarchical levels given the same parameter sizes. BERT\textsubscript{\textit{\fontfamily{ccr}\selectfont tiny}}\_L$8$ represents the model applies BERT\textsubscript{\textit{\fontfamily{ccr}\selectfont tiny}} at the word level and 8 encoder layers at the sentence and document levels. BERT\textsubscript{\textit{\fontfamily{ccr}\selectfont mini}}\_L$1$ represents the model applies BERT\textsubscript{\textit{\fontfamily{ccr}\selectfont mini}} at the word level and only 1 encoder layer at the sentence and document levels. }
\label{tinybertL8_minibertL1}
\renewcommand{\arraystretch}{1.0}
\resizebox{0.8\linewidth}{!}{

\begin{tabular}{cc|cc} 
\hline
         &          & \multicolumn{2}{c}{Hypertension}  \\
         & size{\fontfamily{qcr}\selectfont(M)} & AUC    & PRC                  \\ 
\hline
BERT\textsubscript{\textit{\fontfamily{ccr}\selectfont tiny}}\_L$8$ & {\fontfamily{qcr}\selectfont13.9}     & 0.8684 & 0.8285                 \\
BERT\textsubscript{\textit{\fontfamily{ccr}\selectfont mini}}\_L$1$ & {\fontfamily{qcr}\selectfont13.9}     & \textbf{0.8782} & \textbf{0.8316}                 \\
\hline
\end{tabular}}

\end{table}

\newpage

\subsection{Descriptive statistics of phenotype prediction cohorts}
The MIMIC-III ICD-9 diagnosis table is used to determine phenotypes as the prediction labels. 
The detailed information about phenotypes including disease name and ICD-9 code, and the number of patients from MIMIC-III are shown in Table~\ref{at2}. These are top ten of the most frequent diseases by cumulative patient counts. The selected phenotypes cover the majority of organ systems including circulatory system, genitourinary system, respiratory system, digestive system, and etc. This also 
indicates that our model performs well across a broad spectrum of diseases. 

\begin{table}[h]
\centering
\caption{Descriptive Statistics of Phenotypes}
\label{at2}
\renewcommand{\arraystretch}{1.2}
\resizebox{1.05\linewidth}{!}{%
\begin{tabular}{llll}
\hline
\textbf{Phenotype} & \textbf{ICD-9}& \textbf{Type} & \textbf{\# Patients (\%)} \\ \hline
Essential hypertension & 4019 & chronic & 13399 (43.2) \\

\renewcommand{\arraystretch}{0.9}\begin{tabular}[c]{@{}l@{}}Coronary atherosclerosis\\ of native coronary artery\end{tabular} & 41401 & chronic & 8208 (26.5) \\
Atrial fibrillation & 42731 & mixed & 7525 (24.3) \\
Congestive heart failure & 4280 & mixed & 6473 (20.9) \\
hyperlipidemia & 2724 & chronic & 5387 (17.4) \\
Acute respiratory failure & 51881 & acute & 4329 (14.0) \\
Pure hypercholesterolemia & 2720 & chronic & 3874 (12.5) \\
Esophageal reflux & 53081 & chronic & 3629 (11.7) \\
Pneumonia & 486 & mixed & 2577 (8.3) \\
Chronic airway obstruction & 496 & chronic & 2360 (7.6) \\ \hline
\end{tabular}
}
\end{table}


\clearpage
\subsection{Performance of Different Models on Phenotype Prediction Tasks} \label{detailed values}
We report the performance metrics in AUC (Table~\ref{auc} A), PRC (Table~\ref{auc} B), Precision (Table~\ref{auc} C), Recall (Table~\ref{auc} D), and F1 score (Table~\ref{auc} E) for all phenotype predictions using different models, shown in Table . 

\begin{table}[h]
\centering
\caption{Performance metrics of Different Models for All Phenotypes}

\label{auc}

\renewcommand{\arraystretch}{1.0}
\resizebox{1.0\linewidth}{!}{
\begin{tabular}{lllll}
\multicolumn{5}{c}{\textbf{A. AUC}} \\ \hline
ICD-9 & \textsc{BigBird} & \textsc{Han} & \textsc{BertLstm} & Our Model \\ \hline
4019 & 0.8193 & 0.8331 & 0.8693 & \textbf{0.8720} \\
41401 & 0.8208 & 0.9587 & 0.9482 & \textbf{0.9599} \\
42731 & 0.8023 & 0.9499 & \textbf{0.9565} & 0.9545 \\
4280 & 0.7657 & 0.9075 & \textbf{0.9216} & 0.9212 \\
2724 & 0.7835 & 0.8967 & \textbf{0.9235} & 0.9192 \\
51881 & 0.7424 & \textbf{0.9092} & 0.8902 & 0.9083 \\
2720 & 0.7461 & 0.8044 & 0.6923 & \textbf{0.8693} \\
53081 & 0.7782 & 0.8666 & 0.8882 & \textbf{0.8932} \\
486 & 0.6212 & \textbf{0.8687} & 0.8480 & 0.8666 \\
496 & 0.6178 & 0.8504 & 0.9003 & \textbf{0.9320} \\ \hline
Macro\_AVG & 0.7497 & 0.8845 & 0.8838 & {\color[HTML]{333333} \textbf{0.9096}} \\ \hline
\end{tabular}
}
\end{table}

\begin{table}[h]
\vspace{-0.1cm}
\centering
\renewcommand{\arraystretch}{1.0}
\resizebox{1.0\linewidth}{!}{
\begin{tabular}{lllll}
\multicolumn{5}{c}{\textbf{B. PRC}} \\ \hline
ICD-9 & \textsc{BigBird} & \textsc{Han} & \textsc{BertLstm} & Our Model \\ \hline
4019 & 0.7590 & 0.7817 & 0.8148 & \textbf{0.8166} \\
41401 & 0.6967 & 0.9131 & 0.8938 & \textbf{0.9163} \\
42731 & 0.6589 & 0.8771 & 0.8963 & \textbf{0.8995} \\
4280 & 0.5734 & 0.7592 & \textbf{0.7675} & 0.7665 \\
2724 & 0.4985 & 0.6940 & 0.7309 & \textbf{0.7384} \\
51881 & 0.4068 & 0.6277 & 0.6051 & \textbf{0.6396} \\
2720 & 0.4064 & 0.4522 & 0.2650 & \textbf{0.5594} \\
53081 & 0.4073 & 0.6259 & 0.6532 & \textbf{0.6754} \\
486 & 0.1228 & \textbf{0.4131} & 0.3587 & 0.4084 \\
496 & 0.1167 & 0.4640 & 0.4976 & \textbf{0.6037} \\ \hline
Macro\_AVG & 0.4647 & 0.6608 & 0.6483 & \textbf{0.7024} \\ \hline
\end{tabular}}
 \end{table}

\begin{table}[h]
\centering
\renewcommand{\arraystretch}{1.0}
\resizebox{1.0\linewidth}{!}{
\begin{tabular}{lllll}
\multicolumn{5}{c}{\textbf{C. Precision}} \\ \hline
ICD-9 & \textsc{BigBird} & \textsc{Han} & \textsc{BertLstm} & Our Model \\ \hline
4019 & 0.7187 & 0.7099 & 0.7325 & \textbf{0.7625} \\
41401 & 0.8059 & 0.8644 & 0.8616 & \textbf{0.8775} \\
42731 & 0.8720 & 0.8127 & 0.8456 & \textbf{0.8514} \\
4280 & 0.7403 & \textbf{0.8093} & 0.7605 & 0.7833 \\
2724 & 0.6580 & 0.6642 & 0.6592 & \textbf{0.6667} \\
51881 & 0.6333 & 0.6945 & \textbf{0.6950} & 0.6849 \\
2720 & \textbf{0.7099} & 0.6384 & 0.4043 & 0.5581 \\
53081 & 0.6645 & 0.6779 & \textbf{0.7021} & 0.6467 \\
486 & 0.3684 & 0.5797 & 0.5208 & \textbf{0.5849} \\
496 & 0.3419 & 0.5864 & 0.5301 & \textbf{0.5872} \\ \hline
Macro\_AVG & 0.6513 & \textbf{0.7037} & 0.6712 & 0.7003 \\ \hline
\end{tabular}
}
 \end{table}

\begin{table}[h]
\vspace{-4.5cm}
\centering
\renewcommand{\arraystretch}{1.0}
\resizebox{1.0\linewidth}{!}{
\begin{tabular}{lllll}
\multicolumn{5}{c}{\textbf{D. Recall}} \\ \hline
ICD-9 & \textsc{BigBird} & \textsc{Han} & \textsc{BertLstm} & Our Model \\ \hline
4019 & 0.6769 & 0.7328 & \textbf{0.8155} & 0.7963 \\
41401 & 0.4480 & 0.8055 & 0.7559 & \textbf{0.8176} \\
42731 & 0.3949 & \textbf{0.8305} & 0.8292 & 0.8238 \\
4280 & 0.3713 & 0.5472 & \textbf{0.6762} & 0.5925 \\
2724 & 0.3826 & 0.6536 & \textbf{0.7875} & 0.7821 \\
51881 & 0.3137 & 0.4152 & 0.3913 & \textbf{0.4630} \\
2720 & 0.2861 & 0.3165 & 0.1064 & \textbf{0.6208} \\
53081 & 0.3169 & 0.5924 & 0.6774 & \textbf{0.6979} \\
486 & 0.0279 & \textbf{0.1361} & 0.0850 & 0.1058 \\
496 & 0.2964 & 0.5161 & 0.6083 & \textbf{0.6419} \\ \hline
Macro\_AVG & 0.3515 & 0.5546 & 0.5733 & 0.6342 \\ \hline
\\
\\
\multicolumn{5}{c}{\textbf{E. F1 score}} \\ \hline
ICD-9 & \textsc{BigBird} & \textsc{Han} & \textsc{BertLstm} & Our Model \\ \hline
4019 & 0.6972 & 0.7212 & 0.7717 & \textbf{0.7790} \\
41401 & 0.5759 & 0.8339 & 0.8053 & \textbf{0.8465} \\
42731 & 0.5436 & 0.8215 & 0.8374 & \textbf{0.8374} \\
4280 & 0.4945 & 0.6530 & \textbf{0.7159} & 0.6747 \\
2724 & 0.4838 & 0.6589 & 0.7177 & \textbf{0.7198} \\
51881 & 0.4196 & 0.5197 & 0.5007 & \textbf{0.5525} \\
2720 & 0.4078 & 0.4232 & 0.1685 & \textbf{0.5878} \\
53081 & 0.4292 & 0.6322 & \textbf{0.6896} & 0.6714 \\
486 & 0.0519 & \textbf{0.2204} & 0.1462 & 0.1792 \\
496 & 0.3175 & 0.5490 & 0.5665 & \textbf{0.6133} \\ \hline
Macro\_AVG & 0.4421 & 0.6033 & 0.5919 & 0.6462 \\ \hline
\end{tabular}}
\end{table}


\end{document}